%% file: main.tex
\documentclass[10pt,twocolumn,letterpaper]{article}
\usepackage[pagenumbers]{cvpr}

\input{preamble}
\usepackage{longtable}
\usepackage{needspace}

\definecolor{cvprblue}{rgb}{0.21,0.49,0.74}
\usepackage[breaklinks,colorlinks,allcolors=cvprblue]{hyperref}
\hypersetup{
  pdftitle={Frozen by the Mean: Early Temporal Guidance Unfreezes Motion in Video Diffusion Models},
  pdfauthor={Xuanhua Yin, Boyu Wei, Chengkun Yue, Shunqi Mao, Chuanzhi Xu, Wei Guo, Weidong Cai}
}

\title{Frozen by the Mean: Early Temporal Guidance\\
Unfreezes Motion in Video Diffusion Models}

\input{authors}

\begin{document}
\twocolumn[{%
\renewcommand\twocolumn[1][]{#1}%
\maketitle
\input{figures/teaser}
}]
\AuthorContactFootnote
\input{sec/0_abstract}
\input{sec/1_intro}

\input{sec/2_related}
\input{sec/3_method}
\input{sec/4_experiments}
\input{sec/5_conclusion}
\FloatBarrier
{
    \small
    \bibliographystyle{ieeenat_fullname}
    \bibliography{main}
}

\onecolumn
\appendix
\setcounter{table}{0}
\setcounter{figure}{0}
\setcounter{equation}{0}
\renewcommand{\thetable}{S\arabic{table}}
\renewcommand{\thefigure}{S\arabic{figure}}
\renewcommand{\theequation}{S\arabic{equation}}
\renewcommand{\theHtable}{appendix.\arabic{table}}
\renewcommand{\theHfigure}{appendix.\arabic{figure}}
\renewcommand{\theHequation}{appendix.\arabic{equation}}
\setlength{\LTpre}{8pt}
\setlength{\LTpost}{10pt}
\captionsetup{font=small,labelfont=bf}
\makeatletter
\setlength{\@fptop}{0pt}
\setlength{\@fpsep}{12pt}
\setlength{\@fpbot}{0pt plus 1fil}
\makeatother
\raggedbottom
\section*{Supplementary Material}
\input{sec/supplementary_content}

\end{document}

%% file: preamble.tex
\usepackage[T1]{fontenc}
\usepackage{microtype}

\ifdefined\pdfsuppressptexinfo
\fi

\usepackage{algorithm}
\usepackage{algpseudocode}
\usepackage{array}
\usepackage{tabularx}
\usepackage{colortbl}
\definecolor{thawdark}{HTML}{A63365}
\definecolor{thawlight}{HTML}{FBEFF4}
\makeatletter
\newcommand{\thawrowbackground}{%
  \begingroup\color{thawlight}%
  \smash{\rlap{\rule[-\dp\@arstrutbox]{\linewidth}{\dimexpr\ht\@arstrutbox+\dp\@arstrutbox\relax}}}%
  \endgroup}
\makeatother

\usepackage{placeins}

%% file: authors.tex
\usepackage{fontawesome5}

\author{Xuanhua Yin\textsuperscript{1,\hyperlink{author-contacts}{*}}\quad Boyu Wei\textsuperscript{1}\quad
Chengkun Yue\textsuperscript{2}\quad Shunqi Mao\textsuperscript{1}\\
Chuanzhi Xu\textsuperscript{1}\quad Wei Guo\textsuperscript{1}\quad
Weidong Cai\textsuperscript{1}\\[0.35em]
\textsuperscript{1}The University of Sydney \qquad \textsuperscript{2}Indiana University\\[0.3em]
{\small \href{https://xuanhuayin.github.io/THAW/}{\faGlobe\enspace Project page}
\qquad \href{https://github.com/xuanhuayin/THAW}{\faGithub\enspace GitHub code}}}

\newcommand{\AuthorContactFootnote}{%
  \begingroup
  \renewcommand{\thefootnote}{\fnsymbol{footnote}}%
  \footnotetext[1]{\hypertarget{author-contacts}{}Xuanhua Yin (corresponding author),
    \href{mailto:xuanhua.yin@sydney.edu.au}{xuanhua.yin@sydney.edu.au}.}%
  \endgroup
}

%% file: figures/teaser.tex
\begin{center}
  \captionsetup{type=figure}
  \includegraphics[width=\textwidth]{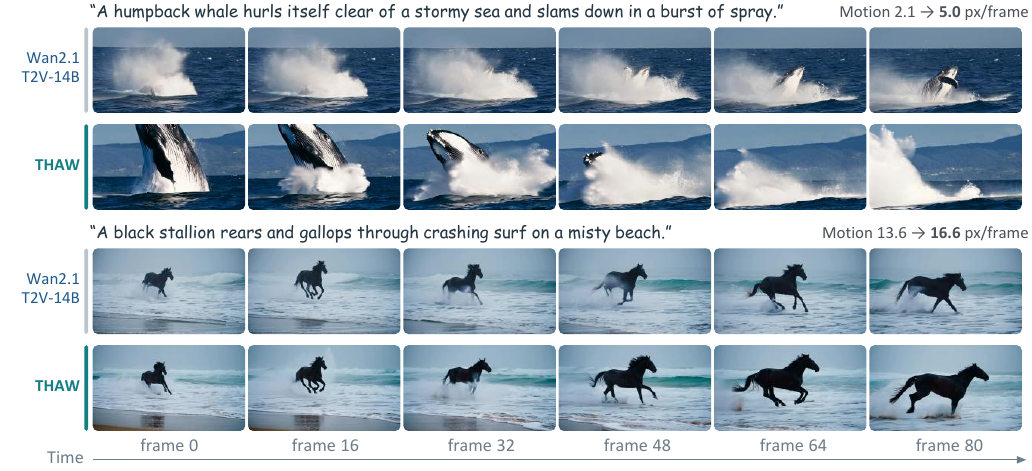}
  \captionof{figure}{\textbf{THAW increases motion without extra model evaluations.}
  Paired Wan2.1-T2V-14B outputs share prompts and sampling budgets (CFG 5, $1280\times720$, 81 frames).
  Frames are time-aligned. Motion is camera-compensated subject flow (px/frame).}
  \label{fig:teaser}
\end{center}

%% file: sec/0_abstract.tex
\begin{abstract}
Video diffusion models can render convincing frames while failing to produce the requested action. Increasing classifier-free guidance (CFG) can strengthen motion, but also saturates and shifts colors. Scalar CFG assigns the same coefficient to the averaged and frame-varying components of its prediction difference, limiting independent control of temporal variation. THAW, Temporal Harmonic Amplification during Warm-up, retains stock CFG and adds a separately weighted residual during an initial denoising window. Its dose specifies the added norm relative to the current CFG difference, giving the extra guidance a consistent relative budget across sampling steps. THAW is training-free and reuses CFG's two model predictions with an unchanged network evaluation count. It improves cost-matched motion over scalar and scheduled CFG on five video models, with positive point estimates against existing guidance methods on three of them. On VBench, THAW raises dynamic degree by about 14 to 15 points over stock CFG while increasing total scores on a different set of three models.
\end{abstract}

%% file: sec/1_intro.tex
\section{Introduction}
\label{sec:introduction}

Video diffusion models generate videos from text descriptions, with an input image providing additional conditioning in image-to-video generation~\cite{ho2022vdm,blattmann2023alignlatents,blattmann2023svd,guo2024animatediff,kong2024hunyuanvideo,wan2025,yang2024cogvideox,ltx2024,hunyuan2025}. Through successive denoising steps, they must produce convincing individual frames and coherent changes over time. Subject motion should remain continuous across frames and convey the actions described by the text prompt.

Even when individual frames look convincing, subject motion can remain weak. Motion-oriented training and preference optimization~\cite{chefer2025videojam,wu2025densedpo,liu2025videoreward} address this issue by updating model parameters. Training-free guidance methods instead change the sampling update~\cite{sadat2024apg,stg2024,shaulov2025flowmo}. Classifier-free guidance (CFG)~\cite{ho2022cfg} steers generation using the difference between two model predictions: one conditioned on the text prompt and the other on an empty or negative prompt. A single scale controls the strength of this guidance. Increasing the scale or varying it over the sampling steps can add motion, but can also alter colors and saturation. The objective is to obtain more motion from these two predictions while limiting faithfulness loss and appearance changes.

Motion involves changes across frames. The CFG prediction difference separates along its frame dimension into an averaged component and a frame-varying residual. For a given pair of predictions, scalar CFG assigns the same coefficient to both components. The residual contains frame-varying responses that affect both motion and appearance. At the same relative norm budget, residual guidance produces similar mean motion gains with smaller saturation changes than guidance along the averaged component. Early residual guidance also produces more motion than late guidance at comparable saturation changes. These motion and saturation differences favor independent residual weighting during early denoising.

We introduce THAW, Temporal Harmonic Amplification during Warm-up, for independent residual guidance. THAW retains the original CFG update and adds a residual term during an initial denoising window. A dose parameter sets the added norm relative to the norm of the CFG prediction difference. Outside this window, sampling uses the original CFG rule. THAW is training-free and reuses the two model predictions required for CFG, keeping the network evaluation count unchanged. Figure~\ref{fig:teaser} illustrates the resulting motion increase.

Cost-matched evaluation measures how much motion a method provides within the same limits on faithfulness and appearance changes. The CFG reference mixes the measured mean motion gains and costs of the tested scalar and scheduled settings. THAW provides more motion than the best cost-matched mixture on five video models. Comparisons with existing guidance methods on three models also yield positive estimates of THAW's motion advantage at matched costs. On VBench~\cite{huang2024vbench}, which evaluates both dynamics and video quality, THAW raises dynamic degree by about 14 to 15 points over stock CFG and improves total scores on a different set of three models.

\input{sec/contributions}

%% file: sec/contributions.tex
\noindent Our contributions are summarized as follows:
\begin{itemize}
  \makeatletter\@clubpenalty=10000\makeatother
  \item We analyze how scalar CFG ties the coefficients of its averaged component and frame-varying residual, and show how direction and timing affect the trade-off between motion and saturation in controlled comparisons.
  \item We introduce THAW, which independently strengthens the residual during early sampling while retaining the original CFG update, without training or additional network evaluations.
  \item We show motion gains beyond the best cost-matched mixtures of the tested scalar and scheduled CFG settings on five video models.
\end{itemize}

%% file: sec/2_related.tex
\section{Related Work}
\label{sec:related}

\subsection{Classifier-Free Guidance}
\label{sec:related_cfg}
Building on classifier guidance~\cite{dhariwal2021beatgans}, CFG controls sampling by scaling the difference between conditional and baseline predictions~\cite{ho2022cfg}, with a predictor-corrector interpretation~\cite{bradley2024predcorr}. Extensions change the baseline prediction~\cite{karras2024autoguidance,ahn2024pag,hong2024seg,sadat2025icg}, decompose update directions~\cite{sadat2024apg,kwon2025tcfg,chung2025cfgpp,saini2025rectcfgpp}, weight spatial frequencies~\cite{sadat2025fdg,lfcfg2025,rychkovskiy2025zeresfdg}, rescale or adapt the guidance strength~\cite{cfgzero2025,lin2024flawed,malarz2025betacfg,wang2026cfgctrl,liu2026trajcfg,gao2026c2fg}, or use past predictions~\cite{sadat2026higs}. These methods provide controls over the strength and geometry of guidance.

\subsection{Motion Control in Video Diffusion}
\label{sec:related_motion}
For video generation, motion must be considered alongside frame quality and temporal consistency, which benchmarks evaluate separately~\cite{huang2024vbench,liao2024devil,ling2025vmbench}. Training-based methods improve motion through objectives or preference data~\cite{chefer2025videojam,wu2025densedpo,liu2025videoreward}. Training-free alternatives modify temporal attention~\cite{bu2024bytheway,luo2025enhanceavideo}, motion concepts in the condition~\cite{kim2026motioncfg} or internal activations~\cite{cheng2026stas}, or intervene on failing samples~\cite{kwon2026earlyfailure}. Other approaches modify the sampling process through noise initialization~\cite{wu2024freeinit,freqprior2025,ge2023pyoco,qiu2024freenoise}, teacher guidance~\cite{lee2025videoguide}, repeated prediction~\cite{jang2026srvs} or filtering of the conditioning image~\cite{alg2025}. FreeInit, VideoGuide and SRVS require extra network evaluations.

\subsection{Temporal Guidance in Video Diffusion}
\label{sec:related_when}
Among sampling methods, temporal guidance is most closely related to our approach. STG constructs a weak prediction by skipping spatiotemporal layers~\cite{stg2024}. FlowMo refines latents using temporal variance~\cite{shaulov2025flowmo}, while TPD restores events in later video frames through a counterfactual prediction and frame- and step-dependent lower bounds~\cite{kang2026tpd}. Guidance timing during denoising is studied through scale schedules~\cite{wang2024cfgsched}, limited guidance intervals~\cite{interval2024} and linear-Gaussian analyses of early guidance~\cite{li2025mechanisms}. THAW decomposes the existing CFG difference across frames and assigns the residual a separate coefficient during early denoising, reusing the predictions already available to CFG.

%% file: sec/3_method.tex
\section{Methodology}
\label{sec:method}

THAW operates inside the sampler of an existing latent video diffusion transformer~\cite{rombach2022ldm,peebles2023dit,esser2024sd3} (Fig.~\ref{fig:method}). It reuses the two CFG predictions, splits their difference across latent frames (Sec.~\ref{sec:projections}), normalizes and adds the residual during an initial window (Sec.~\ref{sec:thaw}), and returns the result to the original scheduler (Sec.~\ref{sec:implementation}).

\input{figures/method}

\subsection{Decomposing Guidance across Frames}
\label{sec:preliminaries}
\label{sec:projections}
\label{sec:diagnosis}
Let $z_i\in\mathbb R^{C\times F\times H\times W}$ be the video latent at denoising step $i$, with $F$ latent frames. The model predicts $v_c=f_\theta(z_i,t_i,c)$ under the text condition and $v_b=f_\theta(z_i,t_i,b)$ under the pipeline's empty or negative condition. In image-to-video sampling, both branches also receive the same conditioning image. These predictions use the model's native diffusion or flow parameterization~\cite{ho2020ddpm,song2021sde,lipman2023flow,liu2023rectflow}. Standard CFG~\cite{ho2022cfg} forms:
\begin{equation}
 g_i=v_c-v_b,\qquad v_{\mathrm{CFG},i}=v_b+w_i g_i.
 \label{eq:cfg}
\end{equation}
For a video-shaped tensor $x$, define the full temporal mean and residual projections by:
\begin{equation}
 \begin{aligned}
 \overline x&=\frac{1}{F}\sum_{f=0}^{F-1}x_f,\\
 P_0x&=(\overline x,\ldots,\overline x),\quad
 P_{\mathrm{AC}}x=x-P_0x.
 \end{aligned}
 \label{eq:fullprojection}
\end{equation}
These are complementary orthogonal projections. The residual contains every nonconstant temporal component, including slow motion, while preserving spatial and channel structure. In \cref{eq:cfg,eq:fullprojection}, scalar CFG multiplies both components by the same $w_i$, so their ratio cannot change at fixed model predictions.

At a fixed step, a frame-shared addition preserves pairwise differences between prediction frames, while a residual addition can change them. The residual also contains frame-varying appearance. Its motion and saturation effects are reported in Sec.~\ref{sec:mechanismresults}.

Several video VAEs encode the first image separately from later temporal groups. For these models, we use the causal projection:
\begin{equation}
 \begin{aligned}
 m&=\frac{1}{F-1}\sum_{f=1}^{F-1}x_f,\\
 \widetilde P_0x&=(x_0,m,\ldots,m),\quad
 \widetilde P_{\mathrm{AC}}x=x-\widetilde P_0x.
 \end{aligned}
 \label{eq:causalprojection}
\end{equation}
The causal residual in \cref{eq:causalprojection} is zero at frame 0 and mean-centered over the remaining latent frames. Below, $P_{\mathrm{AC}}$ denotes the selected full or causal projector.

Keeping the first latent frame separate prevents its distinct encoding from changing the mean subtracted from later frames. The projection acts independently at each spatial position and channel.

The residual energy share quantifies this allocation:
\begin{equation}
 \rho_i=\frac{\|P_{\mathrm{AC}}g_i\|_2^2}{\|g_i\|_2^2}.
 \label{eq:share}
\end{equation}
At the first step on Wan2.1-1.3B, the causal residual holds about $1.6\%$ of the squared norm in \cref{eq:share}. Raising scalar CFG therefore places almost all of its incremental squared norm in the complementary component at that step. The share rises later in denoising (Sec.~\ref{sec:mechanismresults}).

\subsection{Adding a Norm-Controlled Residual}
\label{sec:thaw}
THAW controls the magnitude of the added residual separately from its direction. With a fixed coefficient, the added norm tracks the residual magnitude. Matching the added norm to the full CFG difference gives the dose a common relative-budget interpretation across videos and steps.

Let $\mathcal W=\{0,\ldots,K-1\}$ be the initial denoising window, $d\geq0$ the dose, and $r_i=P_{\mathrm{AC}}g_i$ the residual. For a nondegenerate residual, THAW uses:
\begin{align}
 v_{\mathrm{THAW},i}&=v_{\mathrm{CFG},i}
   +\mathbf1[i\in\mathcal W]\,k_iP_{\mathrm{AC}}g_i,
 \label{eq:thaw}\\*
 k_i&=d\frac{\|g_i\|_2}{\|P_{\mathrm{AC}}g_i\|_2}.
 \label{eq:normalization}
\end{align}
Within $\mathcal W$, the added update has the explicit budget:
\begin{equation}
 \|v_{\mathrm{THAW},i}-v_{\mathrm{CFG},i}\|_2=d\|g_i\|_2.
 \label{eq:normmatch}
\end{equation}

Inside the active window, the effective coefficients are $w_{\mathrm{shared},i}=w_i$ and $w_{\mathrm{AC},i}=w_i+k_i$. At fixed predictions, the complementary projected update is unchanged while the residual gets its own gain. Relative to a same-norm scalar increment $d g_i$, the residual coefficient increment is larger by $k_i/d=\rho_i^{-1/2}$ for $d>0$. At $\rho_i=0.016$, the coefficient is about $8d$ because the residual norm is small, while the added norm remains $d\|g_i\|_2$. Outside $\mathcal W$, THAW uses the original CFG rule.

Early intervention yields more motion at matched saturation change on the evaluated Wan cohorts, even though the residual energy share is larger later in denoising (Sec.~\ref{sec:mechanismresults}). The noise-level analysis below relates early guidance to the final sample.

\subsection{Sampling Algorithm and Model Reuse}
\label{sec:implementation}
Algorithm~\ref{alg:thaw} computes per-video norms over channels, latent frames and space in float32. The extra term is zero when $\|P_{\mathrm{AC}}g_i\|_2\leq10^{-3}\|g_i\|_2$. Both CFG predictions are reused without training, backpropagation or additional network evaluation.

The dose, window and projector are fixed by the protocol in Sec.~\ref{sec:setup}. Motion estimators and visual question answering (VQA) models are used for calibration and evaluation, outside the sampling loop.

\begin{algorithm}[H]
\caption{THAW at Denoising Step $i$}
\label{alg:thaw}
\begin{algorithmic}[1]
\Require $v_c,v_b,w_i,d,\mathcal W,P_{\mathrm{AC}}$
\State $v\gets v_b+w_i(v_c-v_b)$ \Comment{stock CFG}
\If{$i\in\mathcal W$}
  \State $g\gets\operatorname{fp32}(v_c)-\operatorname{fp32}(v_b)$
  \State $r\gets P_{\mathrm{AC}}g$
  \If{$\|r\|_2>10^{-3}\|g\|_2$}
    \State $k\gets d\|g\|_2/\|r\|_2$
    \State $v\gets\operatorname{cast}(\operatorname{fp32}(v)+kr)$
  \EndIf
\EndIf
\State \Return $v$ to the original scheduler
\end{algorithmic}
\end{algorithm}

\subsection{Temporal Response in a Gaussian Model}
\label{sec:interpretation}
An idealized Gaussian model~\cite{li2025mechanisms} relates guidance direction and timing to the temporal response. Let the branches have means $\mu_b,\mu_c$ and a shared covariance eigenbasis, with variances $\lambda_b,\lambda_c$ in a given mode. With exact branch predictions and constant $w$, probability-flow ODE sampling from pure noise to a noise-free endpoint gives guided variance $\lambda_b^{1-w}\lambda_c^w$. Equal branch variances give mean $\mu_b+w(\mu_c-\mu_b)$. For a stationary video Gaussian with circulant temporal covariance, a temporal difference operator $D$ satisfies $DP_0=0$, and:
\begin{equation}
 \mathbb E\|Dx\|_2^2
 =\|D\mu_w\|_2^2+\operatorname{tr}(D\Sigma_wD^\top).
 \label{eq:temporalmoments}
\end{equation}
Here $\mu_w$ and $\Sigma_w$ are the guided mean and covariance. In \cref{eq:temporalmoments}, a frame-shared mean shift contributes no temporal difference. The residual mean and temporal covariance do.

Along $x_t=\alpha_t x_0+\sigma_t\varepsilon$, with $\varepsilon\sim\mathcal N(0,I)$, let $q=\sigma_t/(\alpha_t\sqrt{\lambda})$ for a mode with baseline variance $\lambda$. A constant scalar increment $\delta w$ restricted to $q>q_0$ retains these fractions of the final response to the same increment over the full trajectory:
\begin{equation}
 \begin{aligned}
 R_{\rm mean}(q_0)&=\frac{1}{\sqrt{1+q_0^2}},\\
 R_{\log\operatorname{Var}}(q_0)&=\frac{1}{1+q_0^2}.
 \end{aligned}
 \label{eq:earlyresponse}
\end{equation}
The mean fraction in \cref{eq:earlyresponse} uses equal branch variances. The log-variance fraction is first order in a small branch-variance mismatch.

For $q_0>0$, an early increment retains more of its full-trajectory mean effect than of its first-order log-variance effect under the respective assumptions above. Combined with $DP_0=0$, this distinguishes changes in frame-dependent means from a uniform shift across frames. The supplement gives the integrations.

The variable $q$ measures noise relative to a mode's data variance $\lambda$. A cutoff in $q$ can therefore map to different fractions of steps across modes, models and schedules. THAW uses the target sampler's step coordinates. Section~\ref{sec:efficiency} compares noise schedules at a fixed step count.

%% file: figures/method.tex
\begin{figure*}[t]
  \centering
  \includegraphics[width=\textwidth]{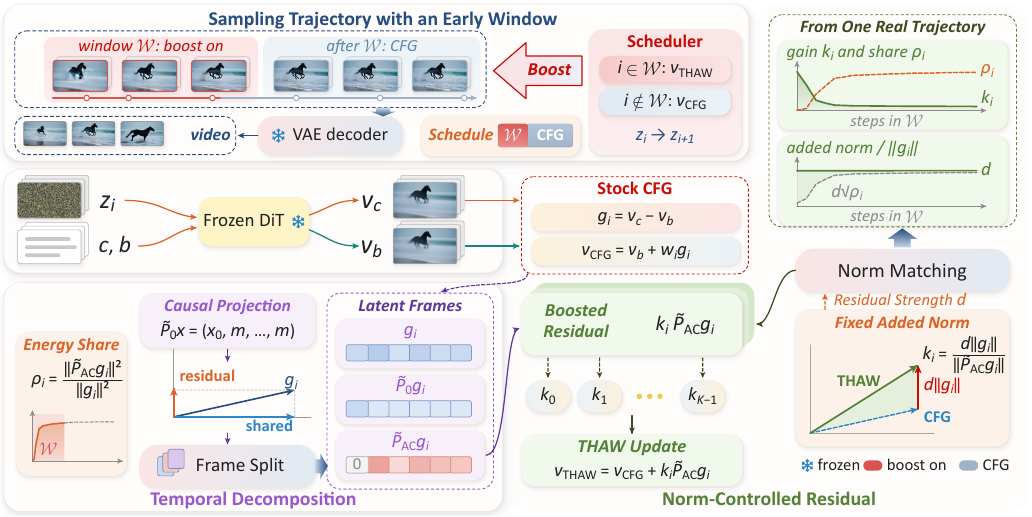}
  \caption{\textbf{THAW adds a norm-controlled temporal residual to CFG during an early window.}
The causal projection splits $g_i=v_c-v_b$ across latent frames. Within $\mathcal W$, THAW adds the residual with gain $k_i=d\|g_i\|_2/\|\widetilde P_{\mathrm{AC}}g_i\|_2$, and after $\mathcal W$ the scheduler uses $v_{\mathrm{CFG}}$. Frames and curves come from one Wan2.1-14B trajectory with $d=2$ over the first 20 of 50 steps. The plotted $\rho_i$ reports residual energy share, and the gray curve shows the added norm from a fixed multiplier $d$.}
  \label{fig:method}
\end{figure*}

%% file: sec/4_experiments.tex
\section{Experiments}
\label{sec:experiments}

The evaluation measures motion at controlled faithfulness and appearance costs, together with the effects of guidance direction, timing and normalization.


\input{figures/experiment_tradeoff}

\subsection{Experimental Setup}
\label{sec:setup}
\label{sec:frontier}
\noindent\textbf{Models and Prompts.}
The evaluated models are Wan2.1-1.3B and 14B~\cite{wan2025}, CogVideoX-5B~\cite{yang2024cogvideox}, non-distilled LTX-Video~\cite{ltx2024}, and HunyuanVideo-1.5~\cite{hunyuan2025}, each with its native scheduler and decoder. Wan uses UniPC~\cite{zhao2023unipc}. Stock CFG is 5 on both Wan models, 6 on CogVideoX and HunyuanVideo, and 3 on LTX. For CogVideoX, LTX and HunyuanVideo, 20 calibration prompts determine THAW's dose, window and projection. Candidate windows cover the first $20\%$, $40\%$ or $60\%$ of the target sampler's steps. Selection maximizes the cost-matched motion gap below among settings with positive motion gain and a matched-motion VQA difference of at least $-0.03$, using calibration point estimates. Both Wan models use the fixed causal setting $d=2$ during the first $40\%$ of steps. Calibration selects $d=3$ with full projection over the first $40\%$ for CogVideoX, and causal $d=4$ over the first $20\%$ and $60\%$ for LTX and HunyuanVideo, respectively. The primary test uses 60 disjoint prompts per model, balanced across animals, objects and humans, with paired initial noise.

\noindent\textbf{Measurements.}
Motion is camera-compensated subject flow $G_{\rm cc}$ (px/frame). RAFT~\cite{raft2020} estimates optical flow, and Grounding DINO~\cite{gdino2023} localizes the subject region. The metric averages residual flow magnitude within the subject region after subtracting flow induced by a background-fitted camera transform. Qwen3-VL-8B-Instruct and InternVL3-8B-hf evaluate 16 uniformly sampled frames per video using normalized yes/no probabilities in $[0,1]$. $M$ averages affirmative-answer probabilities for action and movement questions across both models, and $Q$ averages subject-presence probabilities. Changes $\Delta G$, $\Delta M$ and $\Delta Q$ are paired differences from stock CFG. Appearance costs are relative HSV saturation change $\Delta S$ and mean-color distance $\Delta E$ (CIE76). The four-cost test uses mean absolute saturation change $\mathbb E|\Delta S|$. The motion/saturation plots in Figs.~\ref{fig:tradeoff} and~\ref{fig:mechanism} use $G_{\rm obj}$ and signed saturation.

\input{figures/experiment_qualitative_main}

\noindent\textbf{Four-Cost Test.}
The reference asks how much motion existing CFG settings can provide under THAW's allowed quality costs. The reference includes convex mixtures of the tested stock, scalar and scheduled settings $\mathcal B$: their mean motion gains and costs are combined with nonnegative weights summing to one. Let $\Phi_{\mathcal B}(C)$ be the largest mean motion gain under cost caps $C$. For THAW setting $a$, define $c_a=(-\Delta M_a,-\Delta Q_a,\mathbb E|\Delta S_a|,\Delta E_a)$ and
\begin{equation}
\begin{aligned}
  \Gamma_a&=\Delta G_a-\Phi_{\mathcal B}(C_a),\\
  C_a&=\max(c_a+m,c_0).
\end{aligned}
\label{eq:experimental_gap}
\end{equation}
The first two costs penalize faithfulness losses. The pre-registered tolerances, based on exploratory development measurements, are $m=(0.03,0.03,0.02,1.0)$, with saturation in ratio units. The componentwise maximum keeps the stock point $c_0=(0,0,0,0)$ admissible.

A comparison passes only if the motion gap in \cref{eq:experimental_gap} has a positive one-sided 90\% lower bound LB$_{90}$ and motion VQA satisfies a non-inferiority criterion. The motion-VQA reference $\Psi_{\mathcal B}(\Delta G_a)$ is the largest motion-VQA gain among mixtures reaching at least THAW's motion. When THAW exceeds the baseline motion range, this reference uses the VQA of the maximum-motion baseline point. The criterion is LB$_{90}(\Delta M_a-\Psi_{\mathcal B}(\Delta G_a))>-0.03$. Both tests use 10,000 prompt-bootstrap draws, with Holm correction over the two tests at $\alpha=0.10$ for Tab.~\ref{tab:cross}.

The sixteen existing guidance methods use the same calibration criterion. Of the 48 method-model settings, 31 use pre-registered family defaults because no grid candidate is eligible. Each method comparison adds its selected setting to the CFG reference. Separate joint tests on LTX and CogVideoX include all sixteen selected settings with the CFG controls. Motion-gap tests use Holm correction ($\alpha=0.10$) within pre-registered groups of six, four and six methods on Wan2.1-1.3B, and six and ten on each of LTX and CogVideoX. Motion-VQA non-inferiority is a separate gate using its one-sided 90\% lower bound. Wan2.1-1.3B uses a separate 60-prompt cohort from Tab.~\ref{tab:cross}. LTX and CogVideoX use the same test prompts in both comparisons.

\input{tables/cross_model_main_v3}

\subsection{Main Results}
\label{sec:crossresults}
\label{sec:quality}
\label{sec:i2v}
THAW exceeds the best tested CFG mixture under all four cost caps on five models (Tab.~\ref{tab:cross}). The gain places THAW beyond every admissible tested setting and their mixtures. Repeated-sampling results are in the supplement.

Figure~\ref{fig:tradeoff} shows the saturation cost of increasing flow. ByTheWay's large flow gains accompany large saturation shifts in the 96-prompt sweep (a,b). In the separate 159-prompt comparison (c), scalar scheduling reaches motion similar to low-dose THAW at a much larger saturation increase. Both evaluations use $G_{\rm obj}$ and signed saturation.

\input{tables/literature_main_v3}
\input{figures/experiment_mechanism}

Table~\ref{tab:literature} extends the comparison to existing guidance methods. All 48 cost-matched motion-gap point estimates favor THAW, and 37 comparisons pass the corrected motion-gap test and the motion-VQA gate. Groupwise Holm correction yields no rejections in the ten-method LTX group at $n=60$. The CogVideoX comparison with APG supports a positive motion gap, with inconclusive motion-VQA non-inferiority. The joint sixteen-method motion-gap lower bound is $-0.15$ px/frame on LTX and $+0.01$ px/frame on CogVideoX (one-sided $90\%$). Figure~\ref{fig:qualitative} illustrates subject motion in a separate 100-prompt 14B cohort using fixed baseline parameters transferred from 1.3B. In the displayed case, THAW's kitten follows the rolling yarn with a forward pounce.

VBench evaluates dynamics and quality on 944 prompts, including 72 for dynamic degree, with one video per prompt and method (Tab.~\ref{tab:vbench}). THAW raises total score and dynamic degree over stock CFG on all three models and has the highest dynamic-degree point estimates among the listed methods. The 95\% paired prompt-bootstrap intervals for its dynamic-degree gains are $[4.3,23.2]$, $[6.2,24.7]$ and $[7.2,23.7]$ points on Wan2.1-1.3B, 14B and LTX, respectively (2,000 draws). Imaging quality improves while subject consistency falls on LTX and Wan2.1-1.3B, and motion smoothness decreases on Wan2.1-1.3B. Flicker-consistency scores on VBench's filtered static clips also fall on both models. DEVIL~\cite{liao2024devil} shows a wider dynamics range on 14B with little observed change in controllability.

\input{tables/vbench_main_v4}

Tables~\ref{tab:literature} and~\ref{tab:vbench} hold prompts fixed. The supplement compares THAW with a fixed motion suffix and an LLM rewrite, alone and combined, under the four-cost protocol and on VBench.

\input{tables/human_expanded_main}

Human evaluation compares THAW with scheduled CFG on Wan2.1-1.3B, scalar and scheduled CFG on disjoint LTX pairs, and scalar CFG on HunyuanVideo (Tab.~\ref{tab:humanexpanded}). Preference for more motion and overall preference favor THAW on all three models, with 95\% prompt-bootstrap intervals above 50\%. Only LTX's motion-naturalness interval lies entirely above 50\%, and all color-preference intervals include 50\%.

VMBench~\cite{ling2025vmbench} evaluates perceptible amplitude and temporal quality on a separate 1,050-prompt suite. Against stock CFG, THAW raises amplitude on both Wan models but lowers temporal coherence and motion smoothness. On 14B, the smoothness difference is $-0.033$ (95\% interval $[-0.043,-0.023]$). The increased motion thus carries a temporal-quality cost in this evaluation.

THAW's image-to-video evaluation supplies the same conditioning image to both CFG branches. The custom-prompt cohort uses the first frames of 60 stock-CFG text-to-video outputs as conditions. On Wan2.1-I2V-14B with 40 steps, dose-2 THAW gains $0.127$ px/frame in $G_{\rm obj}$ at $0.4\%$ saturation change, versus ALG's~\cite{alg2025} \mbox{$0.150$ at $6.9\%$}. THAW gives a smaller flow increase with a substantially smaller saturation change on this cohort.

The separate VBench-I2V evaluation~\cite{huang2024vbenchpp} in Table~\ref{tab:i2vmain} uses 246 image-prompt pairs and 50 steps on Wan2.1-I2V-14B-480P at $832\!\times\!480$ with 81 frames. THAW acts during the first $40\%$ of steps, with $d=2$ when combined with ALG. At $d=3$, THAW raises dynamic degree from $23.17$ to $26.83$ relative to stock CFG, while both subject scores decrease by less than $0.3$ points. This setting has higher I2V-subject and subject-consistency scores than ALG, with lower dynamic degree. Adding THAW to ALG raises the combination's dynamic degree and six-dimension average over ALG alone, while lowering both subject scores.

\input{tables/i2v_main}

\subsection{Ablation Studies}
\label{sec:mechanismresults}
Direction, timing and normalization account for the principal design choices. Stock-CFG diagnostics use 60 prompts per curve and one causal projector across models. The residual energy share starts small and grows during denoising (\cref{fig:mechanism}a).

\begin{samepage}
On 100 Wan2.1-14B prompts, averaged (Shared) and residual (THAW) updates use the same relative budget $d\|g_i\|$ along their respective trajectories. Their mean flow gains are close, while the averaged update produces a larger saturation increase (\cref{fig:mechanism}b). Direction therefore affects the saturation cost of increasing flow under this normalization rule. THAW also exceeds temporal noise normalized by the same rule by $0.571$ px/frame (95\% interval $[0.297,0.873]$). The noise comparison supports using the model-derived residual as the added direction.
\end{samepage}

In the 63-prompt Wan2.1-1.3B controls, steps 0 to 2 retain most of the mean flow gain from steps 0 to 11 (\cref{fig:mechanism}c). A separate 96-prompt ablation gives steps 0 to 11 a matched-saturation advantage of $+0.889$ px/frame over late-only steps 12 to 29 (95\% interval $[+0.508,+1.288]$). Earlier intervention gives more flow gain at a similar saturation change, despite the larger residual energy share later.

On the same 96-prompt cohort, RAW2 uses a fixed residual coefficient calibrated to THAW's mean added energy on reference trajectories. The matched-saturation contrast is $+0.296$ px/frame (95\% interval $[-0.011,+0.786]$), whose lower endpoint exceeds the non-inferiority margin of $-0.3$ px/frame. The mean within-prompt flow-gain slope over the tested doses is $0.536$ px/frame per dose (95\% interval $[0.331,0.767]$). These two tests and the early-versus-late test pass Holm correction at a one-sided family-wise level of $0.05$. Norm matching meets the specified non-inferiority criterion, while dose controls the added norm relative to the current CFG difference.

\subsection{Efficiency and Limitations}
\label{sec:efficiency}
THAW reuses stock CFG's predictions and keeps its evaluation count. On 54 Wan2.1-1.3B prompts in the same hardware environment, both use 60 evaluations per video, with measured runtimes of $40.23$ s for THAW and $40.33$ s for stock CFG.

Cost-matched support varies across the evaluated models and noise schedules. Two-expert Wan2.2-A14B yields $\Gamma=+0.25$ px/frame with LB$_{90}=-0.16$. On a separate Wan2.1-1.3B cohort with 30 steps fixed, shift 3 does not pass ($+0.42$, LB$_{90}=-0.36$), while shift 12 passes ($+0.85$, LB$_{90}=+0.17$). At this fixed step count, the statistical decisions differ across flow shifts. A step-based window covers different noise levels under each schedule, so window calibration should use the target sampler.

%% file: figures/experiment_tradeoff.tex
\begin{figure*}[t]
\centering
\includegraphics[width=\textwidth]{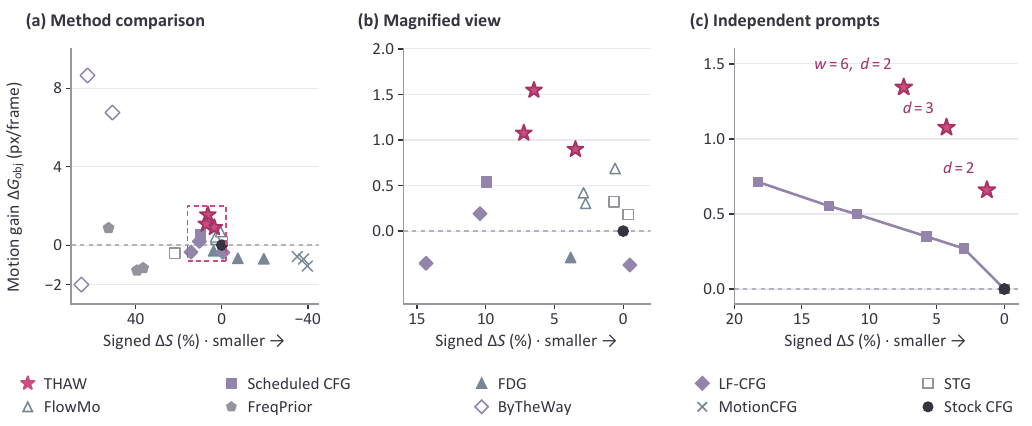}
\caption{\textbf{Trade-offs between motion and saturation on Wan2.1-1.3B.} (a,b) 26 settings on 96 prompts, with a boxed zoom. (c) 159 independent prompts. Upper right means more motion and less saturation increase, with negative $\Delta S$ indicating desaturation. Points are means, and the line connects CFG schedules.}
\label{fig:tradeoff}
\end{figure*}

%% file: figures/experiment_qualitative_main.tex
\begin{figure*}[t]
\centering
\includegraphics[width=\textwidth]{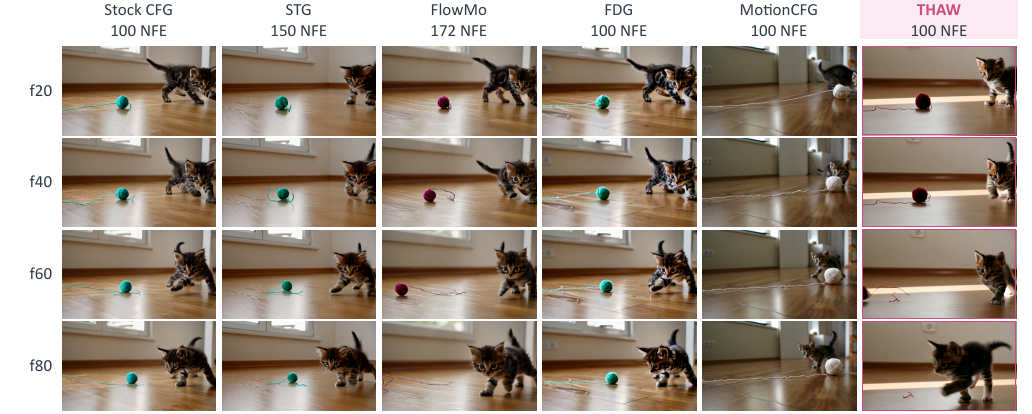}
\caption{\textbf{Qualitative comparison on Wan2.1-14B.} Prompt: ``A kitten chases a ball of yarn across a wooden floor.'' 50 steps, $832\!\times\!480$, 81 frames. Baseline parameters are transferred from the 1.3B model. NFE counts equivalent network evaluations, including FlowMo's backward work.}
\label{fig:qualitative}
\end{figure*}

%% file: figures/experiment_mechanism.tex
\begin{figure*}[t]
\centering
\includegraphics[width=\textwidth]{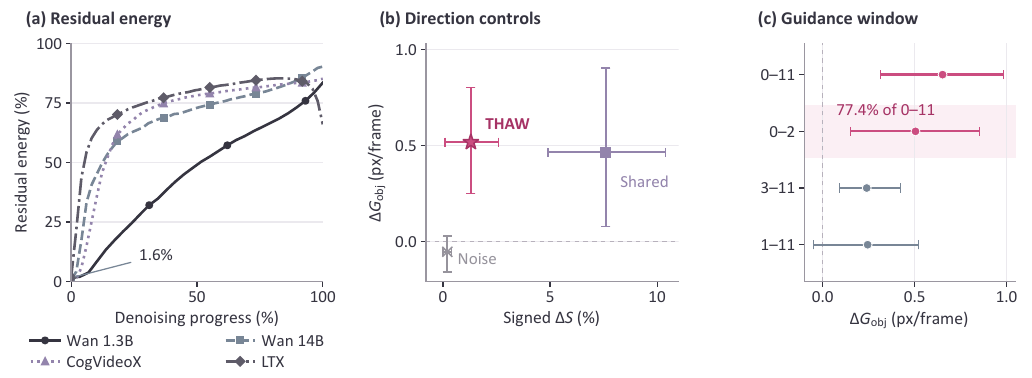}
\caption{\textbf{Direction and timing of residual guidance.} (a) Stock-CFG residual energy share, 60 prompts per curve, with one causal diagnostic projector across models. Progress is sampler step fraction. (b) Wan2.1-14B, 100 prompts: the same relative budget $d\|g_i\|$ on each trajectory. Signed saturation is a relative ratio of cohort means. (c) Wan2.1-1.3B, 63 prompts: windows in zero-based sampler steps. Curves and points are means, bars are 95\% intervals.}
\label{fig:mechanism}
\end{figure*}

%% file: sec/5_conclusion.tex
\section{Conclusion}
THAW gives CFG's frame-varying residual a separate coefficient during early denoising. Its relative-norm budget provides an explicit control over the added guidance while retaining the original CFG update. On five video models, THAW increases motion beyond the cost-matched mixtures of tested scalar and scheduled CFG settings. Comparisons with existing guidance methods, public benchmarks and human preferences characterize the resulting motion and appearance trade-off. The update reuses CFG's predictions and preserves its network evaluation count.

%% file: sec/supplementary_content.tex
\noindent This supplement gives implementation details, evaluation protocols and full experimental results. Section~\ref{supp:implementation} specifies the sampler update. Section~\ref{supp:protocol} defines the comparisons and parameter settings. Sections~\ref{supp:measured}--\ref{supp:extensions} report quantitative results, preferences, controls and extensions. Section~\ref{supp:theory} gives the Gaussian derivations.

\section{Implementation Details}
\label{supp:implementation}

\subsection{Model Predictions and Temporal Projections}
THAW modifies the guided prediction supplied to an existing video sampler. At denoising step $i$, $v_c$ is the conditional prediction and $v_b$ uses the empty or negative condition. Their difference is $g_i=v_c-v_b$, and stock CFG returns $v_{\mathrm{CFG},i}=v_b+w_i g_i$. Both branches receive the same conditioning image in image-to-video sampling. The model's prediction parameterization, scheduler, conditioning embeddings and decoder remain those of the underlying pipeline.

The projection acts on the unpacked $[B,C,F,H,W]$ prediction tensor, where $B$ is batch size, $C$ is the number of latent channels, $F$ is the number of latent video frames, and $H,W$ are spatial dimensions. The full projector subtracts the mean over all $F$ latent frames at each spatial position and channel. The causal projector sets the residual at latent frame 0 to zero and subtracts the mean over frames 1 through $F-1$ from those later frames. Its complementary component retains frame 0 and the mean of the later frames. The full complementary subspace has one temporal degree of freedom at each spatial-channel location, while the causal complementary subspace has two. The causal residual therefore has dimension $F-2$ along the frame axis. The residual includes frame-varying appearance as well as motion.

\subsection{Residual Strength and Numerical Computation}
Let $r_i=P_{\mathrm{AC}}g_i$ for the selected full or causal projector. The residual strength $d\geq0$ controls the added norm during the initial window $\mathcal W=\{0,\ldots,K-1\}$. The guided prediction is $v_{\mathrm{THAW},i}=v_{\mathrm{CFG},i}+\mathbf1[i\in\mathcal W]k_i r_i$, with $k_i=d\|g_i\|_2/\|r_i\|_2$ for a nondegenerate residual. Its allocation can be summarized by the residual squared-norm share:
\begin{equation}
 \rho_i=\frac{\|P_{\mathrm{AC}}g_i\|_2^2}{\|g_i\|_2^2}.
 \label{supp:eq:share}
\end{equation}
For a nondegenerate residual in the active window, the normalization gives:
\begin{equation}
 \|v_{\mathrm{THAW},i}-v_{\mathrm{CFG},i}\|_2=d\|g_i\|_2.
 \label{supp:eq:normmatch}
\end{equation}
For $d>0$, the coefficient increment relative to a same-norm scalar increment is:
\begin{equation}
 \frac{k_i}{d}=\rho_i^{-1/2}.
 \label{eq:allocationgain}
\end{equation}
Thus a small residual norm can produce a large coefficient while the added norm remains $d\|g_i\|_2$. This budget applies to the additional term. The norm of the complete guided prediction can change.

All norms are computed per video over channels, latent frames and spatial positions. The residual calculation uses float32, and the result is cast to the prediction dtype expected by the scheduler. When $\|r_i\|_2\leq10^{-3}\|g_i\|_2$, the extra term is zero. Outside $\mathcal W$, the sampler uses stock CFG. THAW reuses the two branch predictions, with no training, backpropagation or additional model evaluation in the sampling loop. Motion estimators and VQA models are used for calibration and evaluation.

\subsection{Model and Generation Settings}
\Cref{tab:settings} lists resolutions, pixel and latent frame counts, step counts, CFG scales and selected THAW settings. Wan2.1-1.3B and Wan2.1-14B~\cite{wan2025} use checkpoint revisions \texttt{0fad780a} and \texttt{38ec498c}. CogVideoX-5B~\cite{yang2024cogvideox}, LTX-Video 0.9.5~\cite{ltx2024} and HunyuanVideo-1.5~\cite{hunyuan2025} use \texttt{8fc5b281}, \texttt{e58e28c3} and \texttt{286be7ce}, respectively. These abbreviated hashes identify the full revisions in the experiment configurations. Wan2.1 uses UniPC~\cite{zhao2023unipc} and its default negative prompt. CogVideoX, LTX and HunyuanVideo use the empty baseline condition. Paired LTX generations use decoder timestep conditioning $0.05$ and noise scale $0.025$.
\input{tables/supp_settings_v2}

In \texttt{A3w40F}, 3 is the residual strength $d$, 40 denotes the first $40\%$ of denoising steps, and F denotes the full projector. IDs ending in L use the causal projector. We also call $d$ the dose. Let $w_0$ be the stock CFG scale at the current step. \texttt{P1w40} uses $w_0+1$ during its initial window and returns to $w_0$ afterward. \texttt{G1} uses $w_0+1$ at every step. A window of $p\%$ spans the first $K=\operatorname{round}(pN/100)$ of $N$ steps.

Both two-segment CFG schedules switch after the first $40\%$ of steps. \texttt{Tp1m2} uses $w_0+1$ before the switch, and \texttt{Tp2m2} uses $w_0+2$. Both use $w_0-2$ thereafter. In \cref{tab:armsLTX,tab:armsLTXS,tab:qualityLTX,tab:qualityLTXS}, THAW and windowed scalar CFG act for the first $20\%$ of steps, and the two-segment schedules switch at $40\%$.

The additional Wan2.2-A14B experiment uses $832\times480$, 81 frames, 40 steps, flow shift 12, and CFG scales 4/3 for its high-/low-noise experts. Its calibrated THAW setting uses the full projector, strength 3 and the first $40\%$ of steps. Wan2.2-TI2V-5B uses $1280\times704$, 121 frames, 50 UniPC steps, flow shift 5 and CFG 5 in the fixed-transfer experiment. The image-conditioned protocols are specified in \cref{supp:extensions}.

\section{Evaluation Protocols}
\label{supp:protocol}

\subsection{Cohorts, Calibration and Test Sets}
\label{supp:calibration}
The primary cross-model comparison uses 20 calibration prompts and 60 disjoint test prompts per model, balanced across animals, objects and humans. Initial noise is paired across methods. Both Wan2.1 models use the fixed causal setting with strength 2 over the first $40\%$ of steps. For the other models, calibration selects the strength, initial window and projector.

The candidate grid contains causal strengths 2, 3 and 4 with initial windows of $20\%$, $40\%$ and $60\%$, together with strength 1 at $40\%$ and full-projection strength 3 at $40\%$. Eligibility requires a positive motion gain and a matched-motion VQA difference of at least $-0.03$, using calibration point estimates. The selected setting maximizes the admissible-frontier motion gap defined in \cref{supp:fourcost}. Ties prefer lower strength, a window closer to $40\%$, and then the causal projector. The scalar ladder and schedule reference are pre-specified for held-out evaluation. Window calibration uses the target sampler's steps.

The sixteen guidance methods use the same eligibility and selection rule. \Cref{tab:baselinesettings,tab:baselinesettingsLTX,tab:baselinesettingsCOGX} give each selected setting, its candidate grid and its default-fallback status. Of the 48 method-model combinations, 31 use the pre-registered family default because no candidate is eligible. The Wan2.1-1.3B method comparison uses a separate 60-prompt cohort from its primary CFG comparison. LTX and CogVideoX use the same test prompts for both comparisons.
The signed-saturation studies use independent cohorts: a 96-prompt Wan2.1-1.3B method sweep, a 159-prompt Wan2.1-1.3B schedule comparison, and 100 Wan2.1-14B prompts. The 96-prompt ablation cohort is separate from the 96-prompt method sweep. Timing controls use 63 Wan2.1-1.3B prompts. The LTX 49-frame experiment probes clip length, and the calibrated HunyuanVideo experiment supplies a separate model confirmation. Full VBench, VMBench, preference and image-conditioned evaluations use the cohorts identified in their respective subsections.

\subsection{Motion, Faithfulness and Appearance}
The cross-model motion metric $G_{\rm cc}$ is camera-compensated subject flow in px/frame. RAFT estimates optical flow and Grounding DINO localizes the subject~\cite{raft2020,gdino2023}. Background correspondences on an 8-pixel grid support homography RANSAC with 512 hypotheses and a 1-pixel inlier threshold. A 16-pixel dilation excludes the subject from the background fit. The metric averages residual flow magnitude inside the subject region after subtracting the fitted camera flow. When the fitted camera-flow magnitude exceeds twice the observed mean flow plus one pixel, the estimator falls back to affine motion and then median translation. The first temporal seam is excluded from the primary summary. Synthetic camera-only and moving-subject cases test the compensated metric.

The signed-saturation studies report $G_{\rm obj}$, their subject-relative flow metric. It subtracts the vector median of background flow before averaging flow magnitude within the subject mask. The primary summary also excludes the first temporal seam. This median-flow compensation differs from the fitted camera transform used for $G_{\rm cc}$. The two measurements are reported separately.

Qwen3-VL-8B-Instruct and InternVL3-8B-hf evaluate 16 uniformly sampled frames per video using normalized yes/no probabilities in $[0,1]$. Motion VQA $M$ averages affirmative-answer probabilities for action and movement questions across both models. Subject VQA $Q$ averages subject-presence probabilities. For a method $a$ and its paired stock reference $0$, the changes are $\Delta G=G(a)-G(0)$, $\Delta M=M(a)-M(0)$ and $\Delta Q=Q(a)-Q(0)$.

Let $S(a)$ be the video's mean HSV saturation. Signed relative saturation change is $(S(a)-S(0))/S(0)$. The four-cost comparison uses its mean absolute value across videos, $\mathbb E|\Delta S|$, while the signed-saturation frontier uses the signed statistic. Mean-color distance $\Delta E$ is measured in CIE-Lab using CIE76. Saturation is in ratio units in the optimization and in percent where indicated in tables.

\subsection{Four-Cost CFG Mixture Comparison}
\label{supp:fourcost}
Let $\mathcal B$ contain the tested stock, scalar-ladder and selected schedule settings. For setting $j$, $G_j$ denotes its mean motion gain relative to stock CFG, with cost vector:
\begin{equation}
 c_j=(-\Delta M_j,-\Delta Q_j,\mathbb E|\Delta S_j|,\Delta E_j).
\end{equation}
A convex mixture combines these measured mean gains and costs with nonnegative weights that sum to one. The pre-registered tolerances, based on exploratory development measurements, are $m=(0.03,0.03,0.02,1.0)$. For THAW setting $a$, the admissible motion reference is:
\begin{align}
 \Phi_{\rm adm}(a)&=\max_{\lambda\in\Delta_{|\mathcal B|}}\sum_j\lambda_jG_j,\\
 \text{subject to}\quad\sum_j\lambda_jc_{jk}&\leq\max(c_{ak}+m_k,c_{0k})\quad\text{for all }k.
\end{align}
Here $\Delta_{|\mathcal B|}$ is the probability simplex, and the maximum in each constraint is componentwise. The stock cost is $c_0=(0,0,0,0)$, so the stock setting remains admissible. The reported advantage is $\Gamma_{\rm adm}=G_a-\Phi_{\rm adm}(a)$, also denoted $\Gamma$ in the result tables.

For the motion-VQA test, $\Psi(G_a)$ is the largest motion-VQA gain among baseline mixtures with motion gain at least $G_a$. Above the largest tested baseline motion, the reference uses the motion VQA of the maximum-motion baseline point. A comparison passes when the one-sided 90\% lower bound for $\Gamma$ is positive and the lower bound for $\Delta M_a-\Psi(G_a)$ exceeds $-0.03$. The primary CFG comparison uses 10,000 prompt-bootstrap draws and Holm correction over these two tests at $\alpha=0.10$.

Each method comparison adds that method's selected setting to the CFG reference. Its motion-gap tests use Holm correction at $\alpha=0.10$ within the pre-registered groups of six, four and six methods on Wan2.1-1.3B, and six and ten on LTX and CogVideoX. Motion-VQA non-inferiority is a separate gate. Joint references contain ten methods on Wan2.1-1.3B and all sixteen methods on LTX and CogVideoX. \Cref{tab:literaturestats} reports these joint references separately from the individual-method decisions.

The confirmatory four-cost evaluation uses fresh prompts separate from its development experiments. One-sided lower bounds and two-sided $95\%$ intervals are identified separately. Finite degenerate videos remain included with flags and measured or fallback metrics. Exclusion-sensitivity analyses are specific to their source cohorts. Source hashes accompany the numerical results.

\subsection{Signed-Saturation Family Frontiers}
\label{supp:saturation}
For a baseline family $\mathcal F$, let $\Phi_{\mathcal F}$ be its monotone concave motion frontier in signed relative saturation change, including stock CFG and every tested family setting. The family statistic averages over three THAW settings $\mathcal A=\{\mathrm{AC2},\mathrm{AC3},\mathrm{S6AC2}\}$:
\begin{equation}
 \Gamma_{\mathcal F}=\frac{1}{|\mathcal A|}\sum_{a\in\mathcal A}
 \left[\Delta G_{\rm obj}(a)-\Phi_{\mathcal F}(\Delta S(a))\right].
\end{equation}
AC2 and AC3 add causal residuals with strengths 2 and 3 at CFG 5 during the first $40\%$ of steps. S6AC2 additionally raises scalar CFG to 6 within that window. All three use stock CFG 5 outside it. These family-level comparisons use a seven-comparison Holm procedure at $\alpha=0.05$. A raw point in the motion/saturation plot, a single-setting frontier gap and the family-average gap are distinct statistics. The 159-prompt schedule comparison and 100-prompt Wan14B scale comparison use their own paired cohorts and reported intervals.

\subsection{Baseline Settings}
\Cref{tab:baselinesettings,tab:baselinesettingsLTX,tab:baselinesettingsCOGX} report the selected settings, the numbers of candidate and eligible settings, whether the default was used, and the nominal NFE. The nominal budget includes extra evaluations and, where specified, backward or recomputation equivalents. Measured sampling-loop counts are reported separately.
\input{tables/supp_baseline_settings}

\section{Full Results and Video Benchmarks}
\label{supp:measured}

\subsection{Cross-Model Comparisons}
\label{supp:crossmodel}
The primary four-cost comparison supports a positive motion gap over the admissible CFG mixtures on the five main models. The detailed setting tables expose the corresponding motion, camera, saturation, color and VQA changes. The Wan2.2-A14B comparison does not pass the same criterion. Coverage checkmarks indicate comparison availability under the stated protocol. A printed zero is a measured or reference value.
\input{tables/supp_coverage}
\input{tables/supp_COGX}
\input{tables/supp_LTX}
\input{tables/supp_LTXS}
\input{tables/supp_T3}
\input{tables/supp_T2}

The per-method point estimates favor THAW in all 48 comparisons, with 37 passing both the corrected motion-gap test and the motion-VQA gate. Joint comparison changes the conclusion on LTX: its sixteen-method lower bound is negative, while the CogVideoX joint lower bound is positive. The CogVideoX comparison with APG has a positive motion-gap result and inconclusive motion-VQA non-inferiority. These outcomes distinguish motion gain under the cost caps from evidence about motion faithfulness at a comparable motion level.
\input{tables/supp_literature_statistics}

\label{supp:repeated}
Repeated-sampling decisions first average each prompt across the pre-registered runs and then apply the four-cost test. All five primary aggregates pass. The individual-run results are less uniform, with two of the seven additional runs passing on their own. Methods share initial noise within each run. The reference run uses the same initial-noise draw across prompts. Wan2.1-1.3B runs 2 and 3 instead use independent draws for each prompt. The four-run Wan2.1-1.3B aggregate is secondary. \Cref{tab:seedsv3} reports the individual and aggregate outcomes.
\input{tables/supp_seed_v3}

\subsection{Signed-Saturation Comparisons}
The 96-prompt method sweep compares each family's full saturation frontier with the three THAW settings. THAW's mean frontier gap is positive for all seven families and Holm-significant for five. ByTheWay reaches larger flow at larger saturation changes. On the separate 100-prompt Wan2.1-14B cohort, five family comparisons are significant, while the pooled seven-family frontier is not exceeded. The 159-prompt schedule study resolves how scalar and scheduled CFG compare with individual THAW settings at matched saturation. These comparisons use $G_{\rm obj}$ and the signed-saturation protocol of \cref{supp:saturation}.
\input{tables/supp_literature_family_v2}
\input{tables/supp_literature_all}
\input{tables/supp_decisive_frontier}
\input{tables/supp_decisive}
\input{tables/supp_wan14}
\input{tables/supp_wan14_literature}

\subsection{VBench Results}
\label{supp:benchmarks}
VBench~\cite{huang2024vbench} uses its official dimension aggregation on 944 unique prompts, with one video per prompt and method. Each dimension has its own evaluated subset. Dynamic degree uses 72 videos per arm. Temporal flickering is evaluated on filtered static clips, whose valid count can vary by arm. \Cref{tab:vbcounts,tab:vbcontrolcounts,tab:vbflickercounts} record the dimension-specific and static-filtered clip counts. The dynamic-degree differences from stock CFG use 95\% paired prompt-bootstrap intervals with 2,000 draws, as in the main paper. Higher scores are preferred for every VBench dimension. A lower flicker score indicates worse temporal consistency. The headings Object, Multiple, Action and Spatial denote object class, multiple objects, human action and spatial relationship. Appearance, Temporal and Overall denote appearance style, temporal style and overall consistency.

Within each model's paired stock/THAW evaluation, THAW improves dynamic degree on the three main benchmark models. The dimension scores expose accompanying quality changes: subject consistency falls on Wan2.1-1.3B and LTX, and temporal-flickering consistency also decreases on their filtered static clips. The gains in dynamics accompany a loss of temporal quality on these dimensions. Full per-dimension breakdowns for the calibrated literature methods on Wan2.1-1.3B and LTX-Video are reported in \cref{tab:vbmethodsWAN,tab:vbsemanticWAN,tab:vbmethodsLTX,tab:vbsemanticLTX}, while the corresponding Wan2.1-14B evaluations with baseline settings transferred from Wan2.1-1.3B are given in \cref{tab:benchmarktransfer,tab:vbsemanticWAN14}. The separate CFG-control group in \cref{tab:vbreferencecontrols} uses the stock reference shown in that panel.
\input{tables/supp_vbench_methods_WAN}
\input{tables/supp_vbench_semantic_WAN}
\input{tables/supp_vbench_methods_LTX}
\input{tables/supp_vbench_semantic_LTX}
\input{tables/supp_vbench_all}
\input{tables/supp_vbench_counts}
\input{tables/supp_benchmark_transfer}

\subsection{Video Quality on Custom Prompts}
The seven-dimension cross-model evaluation uses the custom 60-prompt cohorts, reusing all 360 videos from each of CogVideoX and the two LTX lengths. Its dimension scores describe these paired CFG comparisons. They supply no full-suite total. On these cohorts, increased dynamics accompanies lower flicker-consistency and motion-smoothness scores.
\input{tables/supp_quality_COGX}
\input{tables/supp_quality_LTX}
\input{tables/supp_quality_LTXS}

\subsection{Additional Benchmarks}
The additional benchmarks separate motion amplitude, temporal quality and dynamic semantics. DEVIL~\cite{liao2024devil} gives Wan2.1-14B a wider dynamics range with little observed change in controllability. T2V-CompBench improves dynamic attribute binding, while its motion- and action-binding intervals include zero. On the independent 1,050-prompt VMBench suite~\cite{ling2025vmbench}, perceptible amplitude rises on both Wan models while object integrity, temporal coherence and motion smoothness fall. The two clip lengths have different scales for accumulated perceptible amplitude.

The 812-prompt VBench-2.0 evaluation improves dynamic attributes and dynamic spatial relationships. Human interaction decreases, while motion-order and motion-rationality point estimates decrease with intervals spanning zero. A separate 98-prompt Wan2.1-14B study distinguishes subject motion from background drift and jitter. An affine camera fit uses background pixels at stride 4 with five Huber-IRLS iterations and a 1-pixel threshold. Subject masks are dilated by $2\%$ of the shorter image side for the background fit. Drift is the mean background residual magnitude, and jitter is the root-mean-square second temporal difference of that residual. Subject flow is measured within the undilated masks. The subject/background measure divides subject flow by background drift plus $0.05$ px.

The paired analysis compares geometric means after metric-specific offsets: $0.01$ for jitter, $0.05$ for drift and subject flow, and $0.01$ for the subject/background measure. Two non-inferiority tests require background jitter and drift ratios below 1.10. A third test requires the subject/background ratio to exceed 1. The three tests use 10,000 paired prompt-bootstrap draws and one-sided Holm correction at $\alpha=0.05$. Both background non-inferiority tests fail, while the subject/background test passes. Background jitter and drift increase alongside the larger relative subject motion. The ratios and model-specific outcomes are in \cref{tab:bigred,tab:extrabenchmarks}. THAW combines positively with STG in the reported 14B comparison, while the FlowMo combination does not show the same gain.
\input{tables/supp_extra_benchmarks_v3}
\input{tables/supp_vmbench_methods}
\input{tables/supp_bigred_v3}

\section{Preference Evaluation and Qualitative Results}
\label{supp:preference}

\subsection{Human Preferences}
Ten participants evaluated 40 fixed video pairs. THAW was compared with scheduled CFG on 16 Wan2.1-1.3B pairs and with scalar CFG on 10 HunyuanVideo-1.5 pairs. The LTX-Video result combines two disjoint groups of seven pairs, one against scheduled CFG and one against scalar CFG. Participants compared the amount of motion, the naturalness of motion and color, and their overall preference. Ties count as half, and 95\% confidence intervals use prompt bootstrapping. All participants gave informed consent.

\Needspace{10\baselineskip}
\input{tables/supp_human_expanded}

In \cref{tab:supphumanexpanded}, the intervals for more motion and overall preference lie above 50\% for all three models. The interval for natural motion lies above 50\% only for LTX-Video. All three natural-color intervals include 50\%.

\subsection{Model-Judged Preferences}
Qwen3-VL-8B and InternVL3-8B judge paired comparisons in both display orders. The reported probability averages valid judges and orders for the same four questions: motion magnitude, natural motion, natural color and overall preference. These model probabilities and the human rating fractions are tabulated independently. In the additional 14B more-motion comparison, both VLM judges favor THAW.
\input{tables/supp_vlm}

\Needspace{25\baselineskip}
\subsection{Paired Video Examples}
\Cref{supp:fig:qualitative} compares THAW and stock CFG with scheduled CFG on LTX-Video and global CFG+1 on CogVideoX.
\input{figures/experiment_qualitative}
The video gallery contains six triplets. Two have positive motion gaps, two are near the median, and two have negative gaps. For each prompt, the gap is measured against the strongest tested scalar or scheduled CFG reference. The figure shows the two positive-gap cases.

The methods share prompts, displayed frame indices, frame coordinates and color rendering. The LTX-Video frames are 0, 40 and 80. The CogVideoX frames are 0, 24 and 48. The 14B comparison in the main paper uses a separate cohort of 100 prompts. Population estimates use the complete cohorts.

\section{Mechanism, Ablations and Efficiency}
\label{supp:mechanism}

\subsection{Direction and Timing}
The stock-CFG diagnostics measure how the squared-norm share in \cref{supp:eq:share} varies during denoising. They summarize up to 60 stored base trajectories per model using the causal-frame-protected projection. First-step shares and averages over the first $40\%$ window answer different questions about the allocation. These model-specific diagnostic cohorts are separate from the held-out comparisons.
\input{tables/supp_spectrum}

The 100-prompt Wan2.1-14B direction controls amplify the averaged component, the temporal residual or temporal noise under the same relative norm rule along each method's own trajectory. Averaged and residual updates give similar mean flow gains, with a larger saturation change for the averaged update. The residual also exceeds matched temporal noise in the paired flow comparison. The HunyuanVideo noise control and its three literature comparisons are not significant. The direction-control evidence is established on the Wan models.
\input{tables/supp_wan14_mechanism}

The 63-prompt timing controls isolate the first few denoising steps within the early window. Steps 0 to 2 retain most of the mean flow gain of steps 0 to 11. A separate 96-prompt ablation compares early steps 0 to 11 with late steps 12 to 29 at matched saturation. Early intervention provides more flow gain at a similar saturation change despite the larger residual energy share later in denoising.
\input{tables/supp_timing_v2}

\subsection{Normalization, Strength and Detailed Ablations}
The 96-prompt ablation cohort covers temporal bands, intervention windows, residual strength, normalization, CFG scale, sampler and clip length. RAW2 uses a fixed residual coefficient calibrated once to THAW's mean added energy on reference trajectories. The normalization contrast tests non-inferiority at a margin of $0.3$ px/frame. The other confirmatory contrasts test the early-window frontier advantage and a positive within-prompt flow-gain slope with strength. All three meet their respective Holm-corrected criteria at a one-sided family-wise level of $0.05$. The individual setting sweeps are descriptive.
\input{tables/supp_ablation_overview}
\input{tables/supp_ablation_confirmatory}
\input{tables/supp_abl_band}
\input{tables/supp_abl_window}
\input{tables/supp_abl_dose}
\input{tables/supp_abl_energy}
\input{tables/supp_abl_cfg}
\input{tables/supp_abl_sampler}
\input{tables/supp_abl_frames}

\subsection{Sampler and Model Transfer}
The additional CFG controls and flow-shift panels use the cohorts and references identified in their captions. At a fixed 30-step count, the tested shift-3 setting does not pass the four-cost criterion while shift 12 passes. A step fraction covers different noise levels under different schedules. The Wan2.2-A14B two-expert setting also fails the four-cost criterion. The 16-prompt $1280\times720$, 121-frame Wan2.1-14B study does not replicate the signed-saturation frontier gain, with a wide interval. The 43-prompt HunyuanVideo comparison with the uncalibrated Wan setting does not confirm that gain. Its conditions differ from the model-calibrated HunyuanVideo comparison.
\input{tables/supp_shift}
\input{tables/supp_cfgcontrols}

The fixed noise-level experiment uses the full projection, strength 3 and a window ending at flow-equivalent noise level $\sigma_f=0.85$. For a variance-preserving schedule, the conversion is $\sigma_f=\sigma/(\sqrt{1-\sigma^2}+\sigma)$. The cutoff gives THAW windows of $26\%$ on CogVideoX, $16\%$ on LTX-Video and $48\%$ on Wan2.2-TI2V-5B. Windowed scalar CFG shares THAW's window on each backbone. \texttt{Tp1m2} switches after 20 of 50 steps in all three panels (\cref{tab:armsCOGXdefault,tab:armsLTXdefault,tab:armsW5default}). CogVideoX and LTX use independent prompts, and Wan2.2-TI2V-5B uses no model-specific calibration.
\input{tables/supp_COGXdefault}
\input{tables/supp_LTXdefault}
\input{tables/supp_W5default}

The SRO comparison in \cref{tab:suppsro} reaches more raw flow together with a large saturation drop. It provides a control with a different motion and appearance trade-off from the residual update.
\input{tables/supp_sro}

\subsection{Sampling Cost}
The paired 54-prompt Wan2.1-1.3B comparison measures THAW and stock CFG in the same hardware environment. THAW keeps stock CFG's network evaluation count. \Cref{tab:runtime,tab:costall} report the measured times and sampling-loop counters. Logged forward counts cover their recorded loop. Nominal whole-method NFE can also include additional conditional passes or backward/recomputation equivalents, so it is listed separately for the literature methods in \cref{tab:baselinesettings,tab:baselinesettingsLTX,tab:baselinesettingsCOGX}.
\input{tables/supp_runtime_v2}
\input{tables/supp_cost}

\section{Conditioning and Prompt Extensions}
\label{supp:extensions}
\label{supp:completion}

\subsection{Image-to-Video Generation}
\label{supp:i2v}
Both CFG branches receive the same conditioning image. The custom-prompt comparison uses the first frames of 60 stock-CFG text-to-video outputs as conditions for Wan2.1-I2V-14B at $832\times480$, 81 frames and 40 steps. It measures $G_{\rm obj}$, saturation, motion VQA and DINO similarity to the conditioning image. Two THAW strengths and two ALG filter strengths~\cite{alg2025} compare additional motion with appearance change and image fidelity. Strength-2 THAW gives a smaller subject-relative flow increase and substantially smaller saturation change than ALG, with a positive gap over the saturation-matched ALG frontier on this cohort.
\input{tables/supp_i2v}

The full VBench-I2V evaluation~\cite{huang2024vbenchpp} is a separate 246 image-prompt-pair benchmark using 50 steps. THAW acts in the first $40\%$ of steps, with strength 2 in its ALG combination. At strength 3, THAW raises dynamic degree while reducing both subject scores by less than $0.3$ points. ALG provides more dynamic degree with lower subject scores at this setting. The THAW+ALG combination raises dynamic degree and the six-dimension average over ALG alone while lowering both subject scores. The six-score average excludes flicker from the seven listed dimensions. These outcomes show how residual strength changes the motion and conditioning trade-off.
\input{tables/supp_i2v_full}

\subsection{Motion Prompting}
\label{supp:prompting}
The fixed motion suffix and the LLM rewrite change the conditioning text while using stock CFG. The four-cost comparison uses the same 20/60-prompt calibration/test design and measures every cost against the original prompt. Motion-gap tests use Holm correction over the two prompt families at $\alpha=0.10$, with a separate motion-VQA non-inferiority gate. On Wan2.1-1.3B and LTX, both edits add more motion than fixed-prompt THAW at about three to four times its saturation change, and THAW passes neither comparison. THAW also adds motion when applied to the rewritten prompt, at a saturation change similar to the LLM rewrite alone.
The LTX fixed-suffix comparison illustrates the two criteria: its positive motion-gap lower bound does not survive Holm correction, and its motion-VQA lower bound is $-0.054$, below $-0.03$.
\input{tables/supp_motion_prompt}

On the VBench prompt suite, both edits raise dynamic degree more than fixed-prompt THAW. The LLM rewrite lowers subject consistency and imaging quality on both models. Both prompt edits also lower those two LTX quality scores. Combining THAW with the rewrite produces the largest dynamic-degree score among these prompt-editing arms on both models. Scoring uses the original prompts for all arms.
\input{tables/supp_motion_prompt_vbench}

\subsection{Negative Conditioning}
\label{supp:negative}
The 60-prompt Wan2.1-1.3B comparison uses the same positive prompts under three baseline branches. The default branch uses Wan's default negative prompt. The neutral branch removes its four motion and stillness items. The empty branch uses an empty negative prompt. Each branch has its own stock-CFG reference and paired changes. THAW passes the within-branch four-cost test under all three conditions.
\input{tables/supp_negative}

\section{Gaussian Temporal Response}
\label{supp:theory}
Linear-Gaussian CFG analysis~\cite{li2025mechanisms} interprets the component-wise response. Consider exact branch predictions with Gaussian data distributions and a shared covariance eigenbasis. For idealized probability-flow ODE sampling from infinite initial noise to zero terminal noise, an eigenmode with conditional and baseline variances $\lambda_c$ and $\lambda_b$ follows the constant-weight CFG variance law $\operatorname{Var}_w=\lambda_b^{\,1-w}\lambda_c^{\,w}$. In the equal-variance special case, covariance stays fixed while the guided mean is $\mu_b+w(\mu_c-\mu_b)$. For a stationary video Gaussian with circulant temporal covariance, the shared temporal eigenbasis also diagonalizes the full temporal projections, and a temporal difference operator $D$ with $DP_0=0$ has
\begin{equation}
 \mathbb E\|Dx\|_2^2=\|D\mu_w\|_2^2+
       \operatorname{tr}(D\Sigma_wD^\top).
 \label{supp:eq:temporalmoments}
\end{equation}
Here $\mu_w$ and $\Sigma_w$ are the guided mean and covariance. With equal branch variances, a constant guidance increment restricted to $q>q_0$ delivers the mean-response fraction $R_{\rm mean}(q_0)=(1+q_0^2)^{-1/2}$. For a small branch-variance mismatch, the first-order fraction of the log-variance response is $R_{\log\operatorname{Var}}(q_0)=(1+q_0^2)^{-1}$. A mode with larger variance becomes sensitive at a larger global noise-to-signal ratio $\sigma/\alpha$.

Consider an affine noising path $x_t=\alpha_t x_0+\sigma_t\varepsilon$ with pure-noise and noise-free endpoints. A shared Gaussian eigenmode has noised variance $V_j(t)=\alpha_t^2\lambda_j+\sigma_t^2$ for branch $j\in\{b,c\}$. The exact probability-flow ODE coefficient multiplying the centered state is
\begin{equation}
 a_w(t)=\frac12\left[(1-w)\frac{d\log V_b(t)}{dt}
                      +w\frac{d\log V_c(t)}{dt}\right].
\end{equation}
Integrating this coefficient from the common pure-noise distribution to the data endpoint gives $\operatorname{Var}_w=\lambda_b^{1-w}\lambda_c^w$, the idealized-endpoint specialization of the Gaussian CFG law~\cite{li2025mechanisms}. This identity supplies the per-mode response used in the temporal interpretation.

For equal branch variances $\lambda$, let $\Delta=\mu_c-\mu_b$ and $q=\sigma/(\alpha\sqrt{\lambda})$. A guidance increment $\delta w$ applied only on the initial interval $q>q_0$ changes the final mean by
\begin{equation}
 \delta m=\delta w\,\Delta\int_{q_0}^{\infty}
          \frac{q\,dq}{(1+q^2)^{3/2}}
       =\frac{\delta w\,\Delta}{\sqrt{1+q_0^2}}.
\end{equation}
The full-trajectory response has $q_0=0$, yielding the mean-response fraction stated above. For a small variance mismatch $r=\lambda_c/\lambda_b$, use $\lambda=\lambda_b$ and expand
\begin{equation}
 \log\frac{V_c(t)}{V_b(t)}
       =\frac{\log r}{1+q(t)^2}+O((\log r)^2).
\end{equation}
Integration over the same initial interval gives
\begin{equation}
 \delta\log\operatorname{Var}
       =\frac{\delta w\log r}{1+q_0^2}
           +O((\log r)^2),
\end{equation}
so the first-order response fraction is $(1+q_0^2)^{-1}$. The temporal specialization assumes a circulant stationary covariance so that its eigenbasis commutes with the full temporal projections. The temporal-difference identity follows from $\mathbb E[xx^\top]=\Sigma+\mu\mu^\top$ and $DP_0=0$.

%% file: figures/experiment_qualitative.tex
\begin{figure}[H]
\centering
\includegraphics[width=\textwidth]{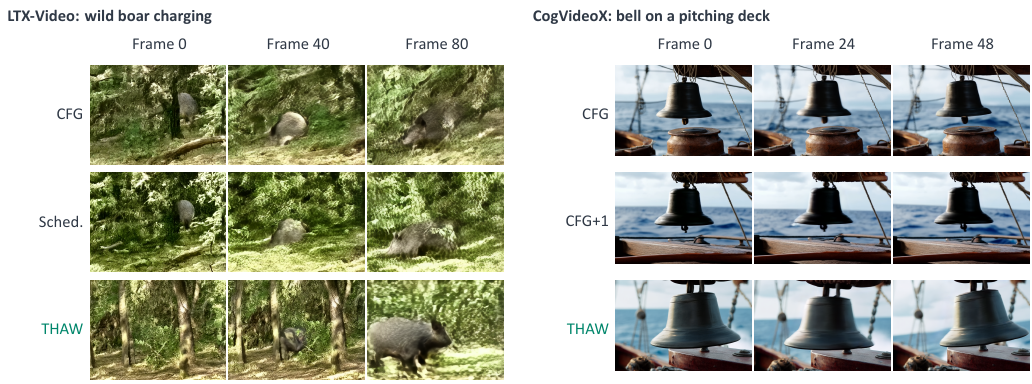}
\caption{\textbf{Paired video generations on LTX-Video and CogVideoX.} Two paired examples: a charging boar (LTX-Video, 161-frame clip) and a swinging bell (CogVideoX, 49-frame clip). Methods share the prompt, checkpoint, scheduler, decoder, and displayed frame indices. Colors and frames are unaltered. The supplementary gallery includes full clips, median cases, and negative measured gains.}
\label{supp:fig:qualitative}
\end{figure}